# IDENTIFICACION Y REGISTRO CATASTRAL DE CUERPOS DE AGUA MEDIANTE TECNICAS DE PROCESAMIENTO DIGITAL DE IMÁGENES EN LANDSAT-5

Kevin H. Rojas Laura  <krojasl@uni.pe>,  Universidad Nacional de Ingeniería, Ing. Electrónica

Christian B. Cárdenas Álvarez  <beni.fc.uni@gmail.com>, Universidad Nacional de Ingeniería, Ing. Física

Asesores:

Dr. Guillermo L. Kemper Vásquez <guillermo.kemper@gmail.com>

Ing. Ernesto Fonseca Salazar <sfonseca@ana.gob.pe>

**Resumen:**

Los efectos del cambio climático mundial sobre los glaciares peruanos han dado origen a varios procesos de desglaciación en los últimos años. El efecto inmediato es la alteración en el tamaño de las lagunas y ríos. Las instituciones estatales que monitorean los recursos hidrológicos actualmente solo tienen estudios recientes sobre menos del 10% del total, los efectos del cambio climático  y la falta de información actualizada acrecentaran los problemas social-económicos relacionados a los recursos hidrológicos en el Perú. El objetivo de este trabajo es desarrollar un aplicativo de software para automatizar el Registro Catastral de Cuerpos de Agua en el Perú, haciendo uso de técnicas de procesamiento digital de imágenes, el cual brindara herramientas para la detección, registro, análisis temporal y visualización de los cuerpos de agua. Las imágenes utilizadas son provenientes del satélite LandSat5, las cuales  pasan por un pre-procesamiento de calibración y corrección propias del satélite, los resultados de la detección, se agrupan en un fichero que contiene los vectores de localización y las imágenes de los cuerpos de agua segmentados.

**Palabras Clave:** LandSat5, Registro Catastral, Cuencas Hidrográficas, Procesamiento de Imágenes Satelitales

**Abstract:**

The effects of global climate change on Peruvian glaciers have brought about several processes of deglaciation during the last few years. The immediate effect is the change of size of lakes and rivers. Public institutions that monitor water resources currently have only recent studies which make up less than 10% of the total. The effects of climate change and the lack of updated information intensify social-economic problems related to water resources in Peru. The objective of this research is to develop a software application to automate the Cadastral Registry of Water Bodies in Peru, using techniques of digital image processing, which would provide tools for detection, record, temporal analysis and visualization of water bodies. The images used are from the satellite Landsat5, which undergo a pre-processing of calibration and correction of the satellite. Detection results are archived into a file that contains location vectors and images of the segmentated bodies of water.

# INTRODUCCIÓN

La gestión de la información es hoy en día la herramienta más valiosa para afrontar los problemas ambientales, las autoridades distritales, regionales y nacionales administradores de los recursos hídricos, necesitan obtener información del comportamiento volumétrico de los recursos hídricos, esto para tener la certeza de tomar la mejor decisión en cuanto distribución, concesión de los recursos sin que afecte al medio ecológico y social.

Debido a los cambios visibles en los volúmenes de nevados que se han originado en la última década (e.g. Nevado del Huascarán, Ancash), ha originado entre muchos otros efectos, el incremento del área de las lagunas y en algunos casos ha dado origen a la creación de lagunas y bofedales estacionales, el monitoreo de estos cambios se realiza solo en un 10% por las instituciones estatales, la Fig.1 muestra las lagunas en estudio por vertiente hidrográfica.

| Vertiente | Total | Lagunas | | | | Lagunas no aprovechadas y sin estudio |
|---|---|---|---|---|---|---|
| | | En explotación 1/ | | En estudio | | |
| | | Número | Capacidad (En millones de metros cúbicos) | Número | Capacidad (En millones de metros cúbicos) | |
| Total | 12 201 | 186 | 3 028 | 342 | 3 953 | 11 673 |
| Pacífico | 3 896 | 105 | 1 379 | 204 | 617 | 3 587 |
| Cerrada | 23 | 3 | 41 | 1 | 185 | 19 |
| Atlántico | 7 441 | 76 | 1 604 | 133 | 3 006 | 7 232 |
| Titicaca | 841 | 2 | 4 | 4 | 145 | 835 |

Figura1 Laguna Según Vertientes, *Fuente:* Sistema Estadístico Nacional Perú Compendio Estadístico. INRENA

El desarrollo de sensores satelitales de gama alta, ha generado que los estudios sobre la tierra y sus recursos naturales se acrecienten, así como también disminuya el tiempo de levantamiento de datos de vastas áreas.

El precio de venta de imágenes satelitales de gran resolución y de software dedicado al Sistema de Información Geográfica (GIS), es la principal barrera para realizar estudios de monitoreo a nivel Nacional en las entidades estatales. Sin embargo hoy en día se cuenta con imágenes de media resolución espacial, disponibles por el Servicio de Geológico de los U.S (USGS), también los entornos de programación han evolucionado, haciéndose más gráficos y sencillos al trabajar con imágenes.

El Programa Espacial LandSat hace posible la descarga de imágenes satelitales desde el servidor Glovis USGS, en él se encuentran imágenes desde el año 1999 hasta la actualidad. Este trabajo hace uso de Imágenes Satelitales de mediana resolución distribuidas desde el servidor Glovis USGS, que sirven para realizar estudios de los cuerpos de agua de tamaño medio.

# OBJETIVOS

El presente trabajo propuesto tiene como finalidad la creación de un aplicativo de software para la detección, medición y análisis de cuerpos de agua a nivel de cuenca, el cual proporcione herramientas específicas a las entidades estatales, regionales y nacionales con el fin que puedan desarrollar un monitoreo permanente sobre una cuenca.

Disminuir el tiempo de adquisición de información catastral de cuencas hidrográficas, solucionando así, un problema de las autoridades administradoras de recursos hídricos en las cuencas del país.

Englobar los conocimientos de Procesamiento de Imágenes, Sistemas Geográficos de Información y programación, orientándolos a solucionar a un problema real de una institución.

# DESCRIPCIÓN DEL PROYECTO

La implementación de los objetivos del proyecto se lleva a cabo en 5 pasos, los cuales se describen en esta sección.
    A. Creación de Base de Datos
    B. Estudio del Satélite
    C. Estudio de los índices

D. Pre procesamiento
E. Procesamiento
F. Desarrollo de Software

A.) Creación de una Base de Datos de Imágenes LandSat5 de la Cuenca a tratar.

Las imágenes del programa satelital LandSat5 están a disposición en sitios webs como www.usgs.gov y www.glcf.umd.edu, la adquisición es creando un usuario y es posible descargar imágenes desde los años 1999 hasta la 2012, con imágenes captadas por el sensor sobre el Perú de cada16 días.

El presente trabajo realiza el estudio sobre una región de interés para la Autoridad Nacional del Agua (ANA) institución nacional administradora de recursos hídricos. Esta región es la Cuenca del Rio Santa, ubicada entre los departamentos Ancash y La Libertad con una extensión de más de 12 mil kilómetros cuadrados.

El área de interés es cubierto con tres imágenes LandSat5, estas son obtenidas desde el servidor Glovis USGS, en secuencias de tres veces por año, en fechas similares, un ejemplo de paquete de datos es mostrado en la Fig. 2. Además es necesario un paquete de imágenes adicionales para hacer pruebas en el software y se obtengan resultados visibles en las primeras pruebas, este paquete de imágenes es la que contiene al lago Chinchaycocha en Junín.

El programa LandSat está compuesto por los últimos sensores TM, TM+, la calidad de imágenes del sensor TM+ fue disminuida desde mayo de 2003, por la falla en el instrumento Scan Line Corrector (SCL-off) debido a esto las imágenes tomadas desde julio de ese año presentan unas franjas de datos erróneas (gaps). Este proyecto hace uso de las imágenes tomadas por el Sensor TM, para evitar el coste computacional de corregir el SCL.

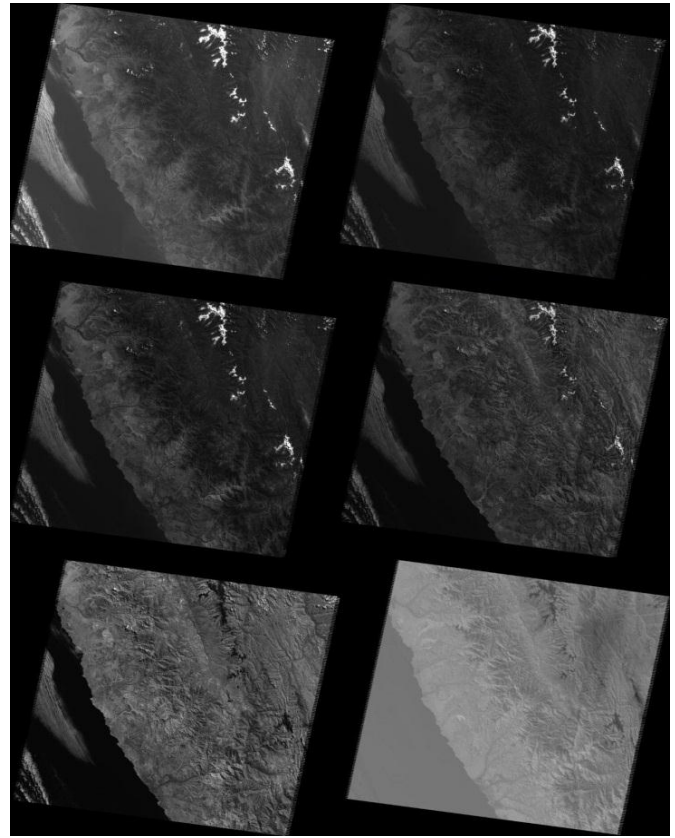

Figura. 2. Paquete de Imágenes Satelitales por Banda del Sensor TM.

Un paquete de datos generado desde el servidor Glovis USGS es mostrado en la Fig. 3, los principales archivos del paquete son las bandas espectrales (e.g LT50070692008122CUB00_B7.TIFF) y la que contiene toda la información del paquete de datos, llamado archivo de Metadatos.

| |
|---|
| **LT50070692008122CUB00_B7.TIFF** |
| **LT50070692008122CUB00_B6.TIFF** |
| **LT50070692008122CUB00_B5.TIFF** |
| **LT50070692008122CUB00_B4.TIFF** |
| **LT50070692008122CUB00_B3.TIFF** |
| **LT50070692008122CUB00_B2.TIFF** |
| **LT50070692008122CUB00_B1.TIFF** |
| **LT50070692008122CUB00_VER.JPG** |
| **LT50070692008122CUB00_VER.TXT** |
| **LT50070692008122CUB00_MTLold.TXT** |
| **LT50070692008122CUB00_MTL.TXT** |
| **LT50070692008122CUB00_GCP.TXT** |
| **README.TXT** |

Figura 3. Ejemplo de contenido de un paquete de datos del servidor Glovis USGS.

B.) Estudio del satélite LandSat5

El satélite LandSat5 es parte de la constelación LANDSAT enviados por los Estados Unidos para el monitoreo de recursos terrestres. El LandSat5 lanzado en 1984 lleva a bordo el sensor Tematic Mapper(TM), un avanzado sensor de barrido multiespectral que opera simultáneamente en siete bandas espectrales, siendo tres en el visible, una en el infrarrojo cercano, dos en el infrarrojo medio y una en el infrarrojo termal.

El estudio del satélite se divide en dos partes, detalles de la características del sensor y análisis de la Metadata que acompaña al paquete de datos del sensor.

B.1) Las características de una imagen Satelital son:

Resolución Espacial:
Este concepto designa al objeto más pequeño que se puede distinguir en la imagen. Está determinada por el tamaño del píxel, medido en metros sobre el terreno.

Resolución Espectral:
Indica el número y anchura de bandas espectrales identificables por el satélite, en general se refiere al número, ancho y espaciamiento de las longitudes de onda a lo largo del espectro electromagnético que el sensor remoto es capaz de identificar.

Resolución Radiométrica:
Se presenta como la capacidad para detectar las variaciones de radiancia espectral, el número máximo de niveles digitales de la imagen suele identificarse como la resolución radiométrica del satélite y esta se indica por el número de niveles de gris captados por el sensor.

Resolucion Temporal:
Es la frecuencia de paso del satélite por un mismo punto de la superficie terrestre. Es decir cada cuanto tiempo pasa el satélite por la misma zona de la Tierra.

Ancho de Barrido:
Es el ancho de la franja de superficie terrestre que es captada por los instrumentos del satélite según pasa por encima.

| RESOLUCIÓN ESPECTRAL |  |
|---|---|
| (micrómetro) | |
| **Banda 1** | 0.45-0.52 |
| **Banda 2** | 0.52-0.60 |
| **Banda 3** | 0.60-0.69 |
| **Banda 4** | 0.76-0.90 |
| **Banda 5** | 1.55-1.75 |
| **Banda 6** | 10.4-12.5 |
| **Banda 7** | 2.08-2.35 |

Figura 4. Resolución Espectral del Sensor TM a bordo del LandSat5.

Es importante conocer los detalles que caracterizan al sensor a utilizar, se resumen las principales características del sensor TM en la Fig. 4 y 5.

B.2) Análisis de la Metadata.

El archivo de nombre (e.g. LT50080662001125CUB00_MTL) de ahora en adelante

MTL, es el archivo que contiene las características propias del sensor y particulares de la imagen satelital. El contenido del archivo Metadata es:

- Información de la Metadata
- Información de la Carpeta
- Producto Metadata
- Atributos de la Imagen
- Valores máximos y mínimos de Radiancias
- Valores máximos y mínimos de ND por banda
- Parámetros de la Proyección sobre el plano

| CARACTERÍSTICA | DESCRIPCIÓN |
|---|---|
| **R. Radiométrica** | 8 bits |
| **R. Temporal** | 16 Días |
| **R. Espacial** | 120 m: banda 6 |
| | 30 m: bandas 1-7 |

Figura 5. Características Radiométrica, temporal y Espacial de LandSat5.

Información de la Carpeta:
Contiene el ID de la imagen LandSat, la fecha y hora de la toma de la imagen y el software usado para su pre procesamiento.

Producto Metadata:
Contiene el nivel de preprocesamiento usado en la corrección de la imagen, el formato de salida de la imagen, el nombre del sensor, la ubicación por en coordenadas sexagesimales y UTM de las esquinas que conforman la imagen y los nombres de los archivos que acompañan a la metadata en la carpeta.

Atributos de la Imagen:
Contiene la cobertura de nubes en el momento de la toma, la calidad de la imagen, el ángulo azimutal y solar del sensor y valores de error calculados en el preprocesamiento de la imagen.

Radiancias:
Valores de Radiancias para cada una de las 7 bandas que conforman la imagen satelital.

Valores máximos y mínimos de ND por banda:
Contiene los máximos y mínimos de los Niveles Digitales por cada banda.

Parámetros de la Proyección sobre el plano:
Contiene los parámetros que caracterizan a la imagen en una proyección, estos son el mapa de proyección, el datum, el elipsoide modelo, la zona UTM y la orientación.

C.) Índices de Vegetación y Agua

La selección de los índices conlleva el análisis previo de cada uno de ellos. El presente trabajo utiliza los índices resultantes de la observación que mejor desempeño muestran, siendo también los que menos costo computacional conllevan.

Los índices de vegetación NDVI se muestra en la expresión (1).

$$\frac{NIR - VIS}{NIR + VIS} \quad (1)$$

Las ecuaciones (2), (3) muestran los índices de agua normalizado y modificado respectivamente a usar en este proyecto.

$$\frac{SWIR - NIR}{SWIR + NIR} \quad (2)$$

$$\frac{VISV - NIR}{VISV + NIR} \quad (3)$$

Donde NIR: near infrared, SWIR: short wavelength infrared y VISV es Visual Verde.

D.) PREPROCESAMIENTO DE LA IMAGEN

Un paso previo a la localización de los cuerpos de agua a estudiar en la imagen satelital, es el calibrado o corrección del sensor, este procedimiento es importante para realizar comparaciones entre imágenes de diferentes fechas, los pasos a seguir son los siguientes.

D.1) Corrección Radiométrica:

El objeto es calcular las Radiancias Espectrales de la imagen Satelital, las variables involucradas en el cálculo están disponibles en el archivo de metadatos de la imagen satelital, en la Fig. 6 se muestran estos para un LandSat5. La ecuación que permite transformar los ND a valores de radiancia ($L$) se puede expresar como:

$$L = L\min + ND*(L\max - L\min)/d\max \quad (4)$$

Donde "dmax" toma el valor de 255 según el MTL.

D.2) Calculo de la Reflectancia Aparente:

Las variables a usarse para el cálculo de la reflectancia se encuentran el archivo de metadatos de la imagen. En este paso usaremos:

- Angulo de elevación solar (e).
- Angulo cenital solar (z=90º - e).
- Día del año juliano (dda).
- Distancia tierra-sol en unidades astronómicas (d).
- Valor medio de la irradiancia solar total en cada banda espectral (E).

La fórmula de cálculo de la reflectancia a partir de la Radiancia está dada por la ecuación (7) ya obtenida en el paso anterior. Un cálculo previo es la obtención de la distancia tierra-sol en el momento de toma, esto es posible mediante la siguiente expresión:

$$d = 1 - (0.01674 \cdot \cos(0.98563 \cdot (dda - 4))) \quad (5)$$

$$\rho = \frac{\pi \cdot d^2 \cdot L}{E \cdot \cos(z)} \quad (6)$$

| PARAMETROS DEL SENSOR TM LANDSAT5 | | | |
|---|---|---|---|
| Banda | Lmin | Lmax | Irradiancia |
| 1 | -1.52 | 193 | 1957 |
| 2 | -2.84 | 365 | 1826 |
| 3 | -1.17 | 264 | 1554 |
| 4 | -1.51 | 221 | 1036 |
| 5 | -0.37 | 30.2 | 215.0 |
| 7 | -0.15 | 16.5 | 80.67 |

Figura 6. Variables definidas para el sensor TM del Landsat5
(Fuente: Nasa, *Landsat 5 Science Data User Handbook.*)

D.3) Corrección Atmosférica:

El uso de las reflectancias para análisis, requiere la conversión de reflectancia aparente a reflectancia de la superficie terrestre, para ello en este trabajo se usa el método desarrollado por Chávez llamado modelo del objeto oscuro.

$$\rho = \frac{\pi \cdot d^2 \cdot (L - L_a)}{E \cdot \cos(z) \cdot t_1 \cdot t_2} \quad (7)$$

Los valores típicos para "t1", "t2" según el trabajo desarrollado por Chávez serian (0.70, 0.78, 0.85 y 0.91) y "cos (z)" respectivamente. En la ecuación 4, la variable "La" es el valor de reflectancia de un pixel cuyo valor teórico de reflectancia es cero.

E.) Procesamiento

En el procesamiento de las imágenes corregidas en la etapa anterior es utilizado conceptos como, los cuales son descritos en esta sección:

- Operaciones Morfológicas
- Binarización

- Representación en falso color

E.1) Operaciones Morfológicas:

La morfología es el estudio de la forma y estructura de los objetos, su estudio es basado en las operaciones de teoría de conjuntos, teoría de retículos, topología y funciones aleatorias. El objetivo de estas operaciones es simplificar y conservar las principales características de forma de las regiones. Las operaciones morfológicas usadas por el software son la dilatación y erosión. Estas son descritas a continuación.

- Dilatación: La dilatación se describe como un crecimiento de pixeles, es decir, se marca con 1 la parte del fondo de la imagen que toque un pixel que forma parte de la región. Esto permite que aumente un pixel alrededor de la circunferencia de cada región y así poder incrementar dimensiones, lo cual ayuda a rellenar hoyos dentro de la región.

$$A \oplus B = \langle x \in X | x = a + b; a \in A, b \in B \rangle$$

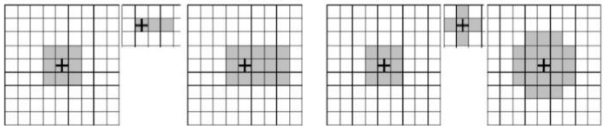

Figura 7. Ejemplo de la operación morfológica Dilatar.

- Erosión: La erosión de A por B se puede entender como el lugar geométrico de los puntos alcanzados por el centro de B cuando B se mueve dentro de A. Un ejemplo se muestra en la Fig.8.

$$A \ominus B = \langle x \in X | x + b \in A; \forall b \in B \rangle$$

La aplicaciones de operadores morfológicos en el software, es para la obtención de características de la imagen como el área y el perímetro, para ello es necesaria la aplicación sucesiva de la dilatación y erosión con el fin de obtener los bordes de la imagen segmentada. El diagrama de bloques se explica mejor en la sección de Desarrollo de Software.

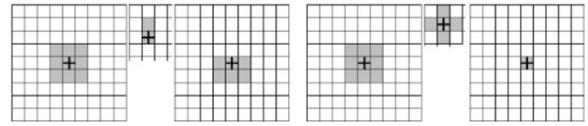

Figura 8. Ejemplo de la operación morfológica Erosionar

E.2) Binarizacion

Consiste en un proceso de reducción de la información de una imagen digital a dos valores: 0 (negro) y 255(blanco). Esta técnica consiste en comparar cada pixel de la imagen con un determinado umbral (valor límite que determina si un pixel será de color blanco o negro). Los valores de la imagen que sean mayores que el umbral toman un valor 255 (blanco), el resto de pixeles toman valor 0(negro). En la Figura siguiente muestra un ejemplo de binarizar una imagen digital.

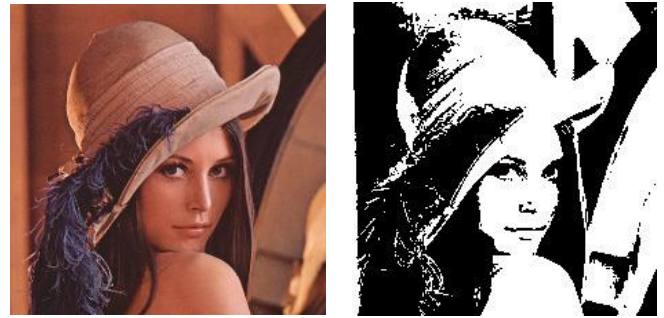

Figura 9. Imagen Digital binarizada con un umbral de 95.

E.3) Representación en Falso Color.

La visualización en falso color implica hacer una combinación de 3 bandas similar a la que se hizo en color verdadero RGB, sin embargo en este caso se combinan bandas que no necesariamente pertenecen al espectro visible con el fin de visualizar características resaltantes de la combinación de las bandas. Un ejemplo para visualizar en falso color es colocar la banda 5 como n=3, banda 4 como n=2 y banda 3 como n=1. El resultado es mostrado en la Fig. 10.

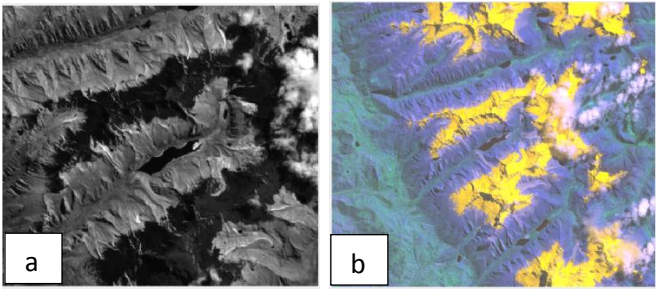

Figura 10. Representación de bandas en falso color (3,4,5), (a) es la banda 5 de imagen y (b) es la imagen a falso color.

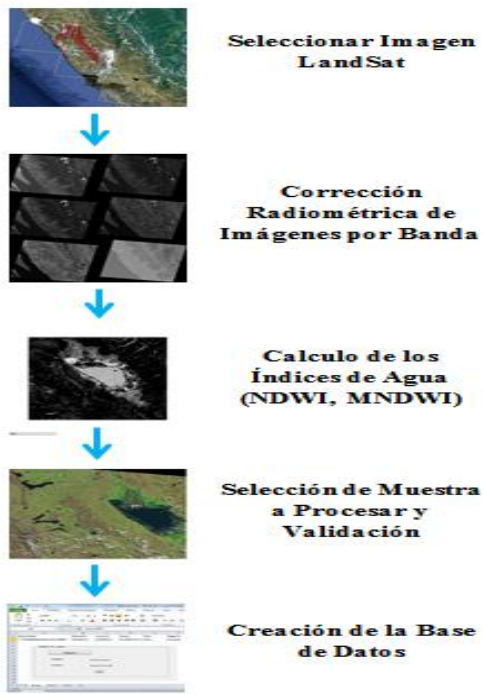

Figura 11. Descripción Grafica del Procedimiento Usado en la interfaz de usuario.

F.) Desarrollo de Software.

La Interfaz Gráfica de Usuario (GUI) es divido en zonas, cada una de estas zonas es conectado con las demás por medio de botones y eventos que llaman a resultados calculados por otras zonas de la interfaz, esto sugiere un procedimiento en procesar, segmentar y registrar, la Fig.12 muestra el esquema de la GUI. La descripción del procedimiento general para la interfaz es mostrado en la Fig. 11. A continuación se describen los pasos a seguir.

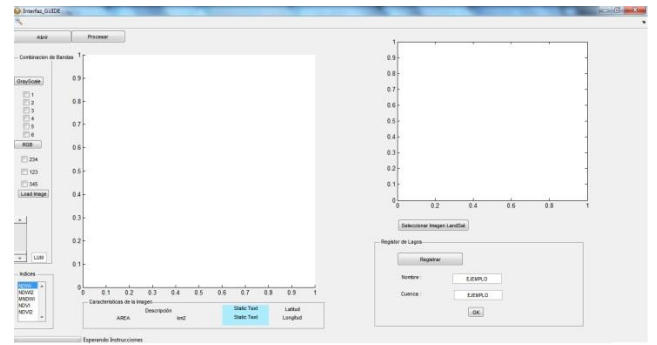

Figura 12. Esquema General de la Interfaz Grafica de Usuario(GUI).

F.1) Seleccionar Imagen LandSat:

La secuencia de pasos para la obtención de información de las lagunas empieza por la adquisición de las imágenes de una base de datos ya creada en un fichero del computador, esta base de datos es obtenida desde el servidor Glovis USGS. Un ejemplo de la base de datos de imágenes es mostrado en la Fig.13. La región de interés(ROI) está conformada por 3 paquetes de imágenes LandSat5; la Interfaz Gráfica de Usuario(GUI) muestra la opción de elección de una imagen que es parte de la ROI, una vez seleccionada se importan automáticamente todas las características del archivo Metadata y los mapa de bits de las 7 bandas espectrales de la imagen seleccionada.

F.2) Corrección Radiométrica de Imágenes por Banda:

Las correcciones necesarias para las 6 bandas son realizadas según las ecuaciones (4), (5), (6), (7), los parámetros requeridos se encuentran en los datos importados desde el archivo Metadata.

F.3) Calculo de los índices de Agua.

La GUI tiene una lista de índices, el usuario realizara una selección del índice, con el cual se ejecutara la segmentación de lagunas. El cálculo de los índices es llevado a cabo usando las 6 bandas corregidas en el paso anterior. La Fig. 14 los índices calculados para una región.

Figura 13. Conjunto de Imágenes LandSat5, clasificadas por fecha y lugar de adquisición.

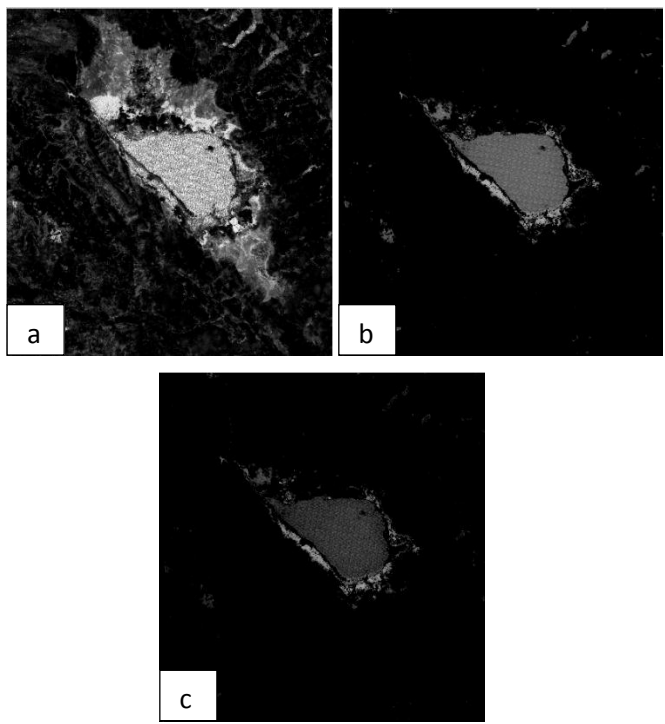

Figura 14. Índices de NDWI (a), MNDWI (b) y NDVI(c) calculados sobre la laguna Chinchaycocha en Junín.

F.4) Selección de Muestra a Procesar y Obtención de Características de la Imagen.

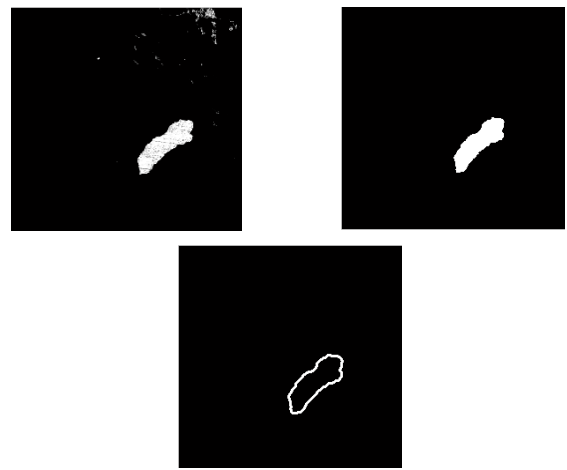

Figura 15. Obtención de bordes de la laguna con operadores morfológicos.

La selección de un punto de referencia dado por el usuario, inicia una secuencia interna de procesamiento de una zona cuyo centro es el punto de referencia dado por el usuario, el resultado son los parámetros a utilizar para la binarización de la imagen, el proceso se basa en el estudio del histograma de la zona seleccionada. Al finalizar la binarizacion se tiene solo la laguna más cercana al punto de referencia que cumple con los parámetros resultantes del estudio del histograma.

La obtención de los bordes de la laguna segmentada es realizada con la aplicación de operadores morfológicos sobre la imagen. En la Fig. 15 y se muestra el proceso llevado a cabo, además en el anexo se adjunta el código que lo ejecuta.

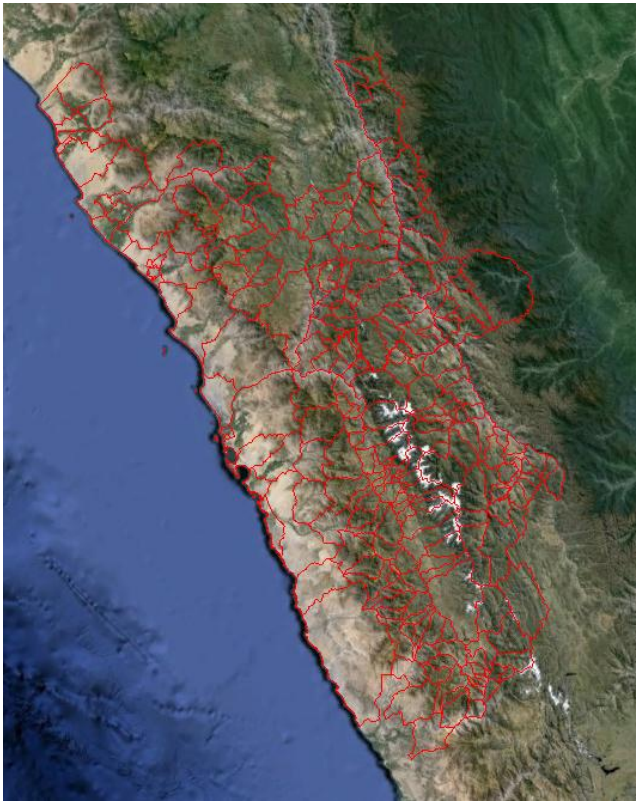

Figura 16.Vector de Coordenadas de los Distritos de Ancash y La Libertad visualizadas en Google Earth. *Fuente: ANA*

F.5) Creación de la Base de Datos

Con la validación hecha sobre la zona de laguna seleccionada por el usuario, sigue la creación de una tabla con las características de la laguna, un dato importante es la localización de la laguna en el mapa del Perú. Para esto importamos un archivo "Mapa_Distrital_Peru.shp" con los vectores que componen la frontera de los distritos de todo el Perú, archivo brindado por el ANA, la Fig.16 muestra las coordenadas de las fronteras de los distritos de Ancash y La Libertad. Haciendo uso de la función conversor de UTM a sexagesimales (función implementada adjuntada en los anexos), las coordenadas del centroide son guardadas en sexagesimales. Calcular la región, provincia, distrito a la que pertenece el centroide de la laguna con las funciones implementadas, para luego guardarlas en la Base de Datos.

El cálculo del área de laguna es realizado con una función implementada que realiza la equivalencia pixel-metro, el valor del área es guardado en la base de datos usando unidades de km.

## RESULTADOS

El aplicativo de software con las herramientas para realizar detección, segmentación y creación de base de datos de lagunas sobre la cuenca del Rio Santa es finalizado en su etapa de prueba. La Fig. 17 muestra el aplicativo de software en funcionamiento sobre la cuenca del rio Santa en Ancash.

Las primeras pruebas sobre el aplicativo de software se realizaron usando imágenes del año 2007, en esas imágenes se lograron segmentar más de 10 lagunas de considerable tamaño. La Fig.13 muestra la carpeta con las imágenes resultantes de estas primeras pruebas.

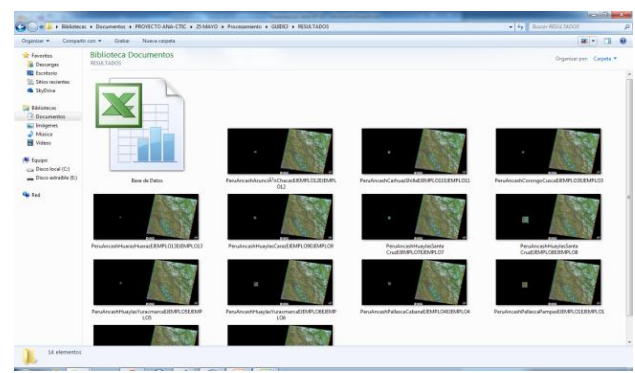

Figura 18. Lagunas segmentadas junto con la imagen LandSat5 que le dio origen.

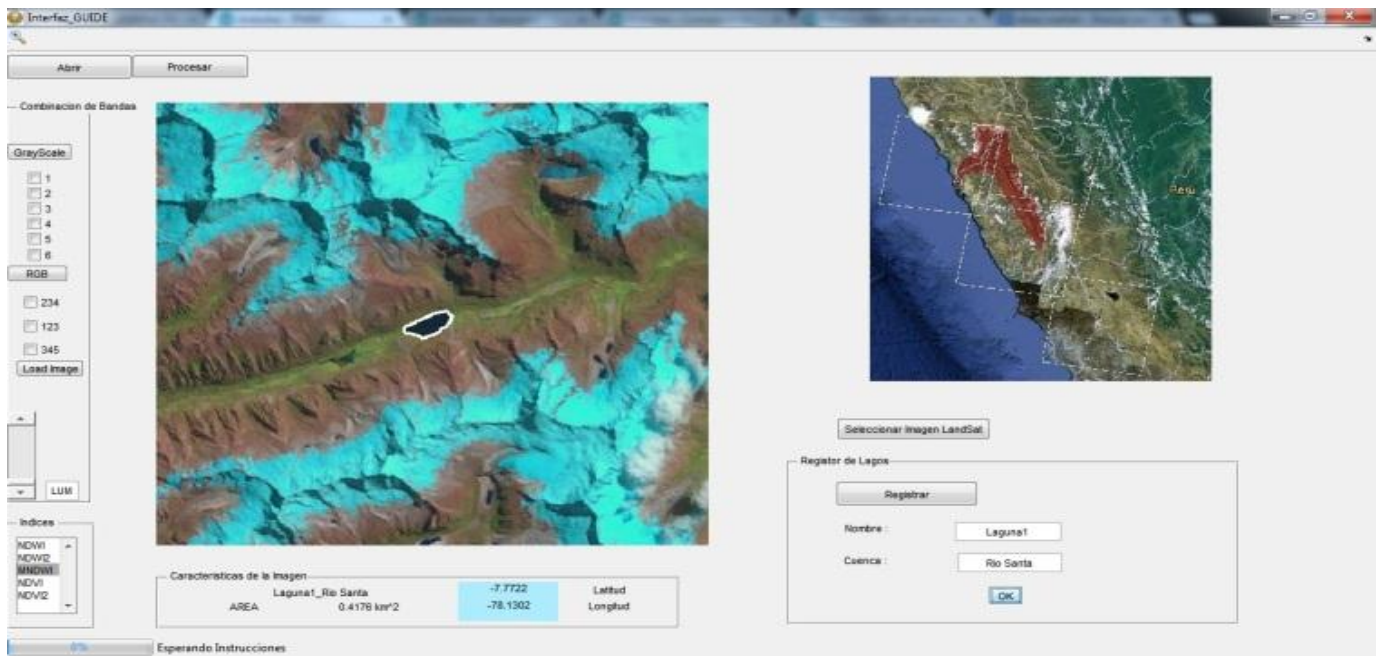

Figura 17. Resultado de segmentación de una laguna usando el aplicativo de software en la Cuenca del rio Santa.

Los resultados de análisis temporal para el proyecto, se basan en el estudio de tres lagunas en el departamento de Ancash (Ver anexo1), las lagunas son:

- Laguna Pelagatos
- Laguna Paron
- Laguna Qerocha

Los paquetes de imágenes fueron adquiridas con una frecuencia de una vez por año, durante los años 2007, 2009, 2011 desde servidor Glovis USGS. Las lagunas son seleccionadas usando el aplicativo de software, las características resultantes como ubicación, área, perímetro son guardadas junto con la imagen de la laguna segmentada, en las Fig.19 y 20 se muestra las características calculadas para las tres lagunas por año.

| 2007 | | | | |
|---|---|---|---|---|
| IDLandSat | Año | Nombre | Cuenca | Area km^2 |
| "LT50080662008129CUB0 | 2008 | Pelagatos | Santa | 1.7739 |
| "LT50080662008129CUB0 | 2008 | Paron | Santa | 1.4724 |
| "LT50080672007142CUB0 | 2007 | Querocha | Santa | 1.3851 |
| 2009 | | | | |
| IDLandSat | Año | Nombre | Cuenca | Area km^2 |
| "LT50080662009179CUB0 | 2009 | Pelagatos | Santa | 1.9953 |
| "LT50080662009179CUB0 | 2009 | Paron | Santa | 1.6947 |
| "LT50080672009131CUB0 | 2009 | Querocha | Santa | 1.4112 |
| 2011 | | | | |
| IDLandSat | Año | Nombre | Cuenca | Area km^2 |
| "LT50080662011137CUB0 | 2011 | Pelagatos | Santa | 1.7667 |
| "LT50080662011137CUB0 | 2011 | Paron | Santa | 1.494 |
| "LT50080672011153CUB0 | 2011 | Querocha | Santa | 1.4067 |

Figura 19. Áreas calculadas para las lagunas Pelagatos, Paron y Querocha durante los años 2007, 2009 y 2011.

| 2007 | | | | | |
|---|---|---|---|---|---|
| IDLandSat | Año | Nombre | Cuenca | Centroide Latitud | Centroide Longtiud |
| "LT5008066200 8129CUB00" | 2008 | Pelagatos | Santa | -8.179486595 | -77.79499326 |
| "LT5008066200 8129CUB00" | 2008 | Paron | Santa | -8.993284653 | -77.66900783 |
| "LT5008067200 7142CUB00" | 2007 | Querocha | Santa | -9.717370767 | -77.32451465 |
| 2009 | | | | | |
| IDLandSat | Año | Nombre | Cuenca | Centroide Latitud | Centroide Longtiud |
| "LT5008066200 9179CUB00" | 2009 | Pelagatos | Santa | -8.179496006 | -77.79363317 |
| "LT5008066200 9179CUB00" | 2009 | Paron | Santa | -8.992750445 | -77.66791335 |
| "LT5008067200 9131CUB00" | 2009 | Querocha | Santa | -9.717370767 | -77.32451465 |
| 2011 | | | | | |
| IDLandSat | | Nombre | Cuenca | Centroide Latitud | Centroide Longtiud |
| "LT5008066201 1137CUB00" | 2011 | Pelagatos | Santa | -8.179486595 | -77.79499326 |
| "LT5008066201 1137CUB00" | 2011 | Paron | Santa | -8.993015575 | -77.66873322 |
| "LT5008067201 1153CUB00" | 2011 | Querocha | Santa | -9.717370767 | -77.32451465 |

Figura 20. Coordenadas de Centroides calculados para las lagunas Pelagatos, Parón y Querocha durante los años 2007, 2009 y 2011.

La ubicación geográfica de los centroides de las lagunas generadas por el software, son guardadas con una extensión compatible a Google Earth; la Fig.21 muestra los resultados de ubicación de las lagunas.

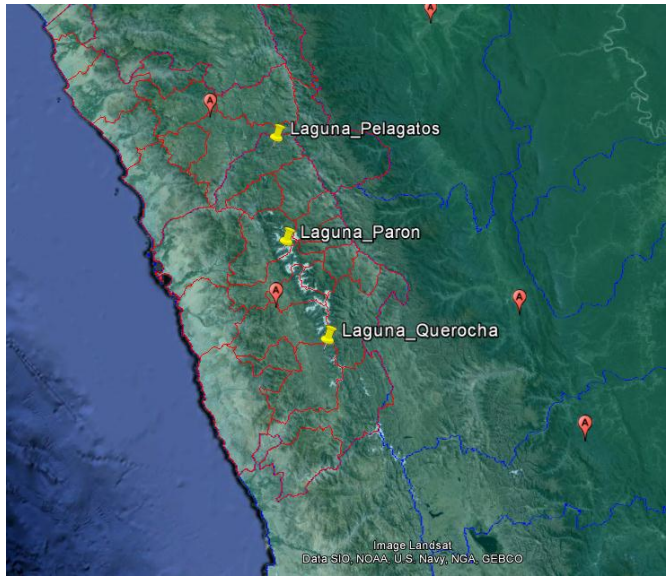

Figura 21. Ubicación resultante de las lagunas Pelagatos, Paron y Querocha visualizado en Google Earth.

La visualización de los resultados generados es posible con la superposición de los bordes calculados por el software en la imagen a color de las lagunas, la Fig.22 muestra las imágenes de lagunas superpuestas con sus respectivos bordes.

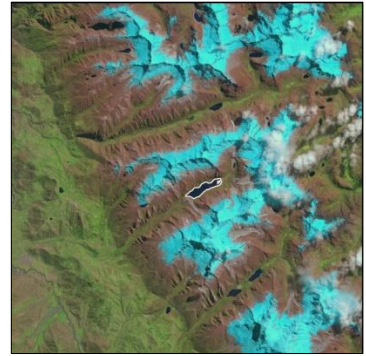
(a)

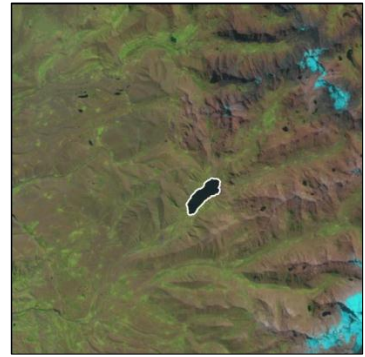
(b)

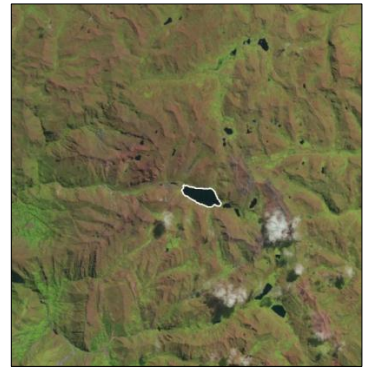
(c)

Figura 22. Imágenes a en falso color de las lagunas Pelagatos(a), Parón (b) y Querocha(c), resaltadas por la segmentación de sus bordes.

**CONCLUSIONES**

La obtención de características como área, perímetro, ubicación geográfica e imagen segmentada, es realizada automáticamente por el aplicativo de software propuesto.

Acelerando el levantamiento de datos sobre las cuencas hidrográficas del país.

Las imágenes satelitales es una fuente de información abundante de los recursos naturales; los gobiernos y autoridades deben extender su uso y aplicación en el monitoreo, estudio y administración de recursos.

Es posible realizar un monitoreo remoto a resolución media de las cuencas hidrográficas para las instituciones estatales utilizando imágenes satelitales libres.

## RECOMENDACIONES

El procedimiento de calibrado de una imagen, requiere de una corrección geométrica, en un futuro proyecto se adicionara las correcciones al software, con la finalidad de disminuir el error en los resultados obtenidos.

## REFERENCIAS BIBLIOGRÁFICAS

Fabián Reuter, "Plataformas Orbitales y Sensores". Series Didácticas N° 34, Cátedra de Teledetección y Cartografía. Universidad Nacional de Santiago del Estero.

Emilio Chuvieco y Stijn Hantson, "Plan Nacional de Teledetección de Media Resolución Procesamiento estándar de imágenes Landsat, Documento técnico de algoritmos a aplicar" (2010), Universidad de Alcalá.

Chavez, P. S. (1996). "Image-based atmospheric corrections. Revisited and improved." Photogrammetric Engineering and Remote Sensing 62(9): 1025-1036.

Chander, G., B. L. Markham, et al. (2007). "Revised Landsat-5 Thematic Mapper Radiometric Calibration." IEEE GEOSCIENCE AND REMOTE SENSING LETTERS 4(3): 490-494.

Aplicación de métodos de corrección atmosférica de datos Landsat5 para análisis multitemporal, Brizuela, Armando B.; Aguirre, César A.; Velasco, Inés

PERÚ. Instituto Nacional de Estadística e Informática. *PERÚ, COMPENDIO ESTADÍSTICO 2008*. Lima, Perú. Gobierno del Perú, 2008.

Tulio Chávez; Colonia Ortiz; Loarte Cadenas; Albornoz Albornoz y Alex Zambrano Ramírez. (2011) "Identificación de lagunas de Alta Montaña en el Perú mediante técnicas de Teledetección Espacial y Modelos de Elevación Digital". ANAIS XV SIMPÓSIO BRASILEIRO DE SENSORIAMENTO REMOTO – SBSR.

Cem Unsalan and Kim L. Boyer. Multispectral Satellite Image Understanding, Springer, 2011.

**ANEXOS**

Función UTM2Geod en MatLab
```
function [ Latitud Longitud ] = UTM2Geod( X, Y, huso, Hemisferio )
% Conversor de UTM a Sexadeciaml.
% Entradas X Y Norte o Sur.
% Uso de las ecuaciones de Coticchia-Surace.
% Si la entrada Y<0 entonces se lesuma 10000000.
if Y <0
    Y = Y + 10000000;
end
% Semieje Mayor
EjeA = 6378388.0;
% Semieje Menor
EjeB = 6356911.946130;
% Excentricidad
e = sqrt((EjeA)^2- (EjeB)^2)/(EjeA);
eprima = (sqrt((EjeA)^2- (EjeB)^2)/(EjeB));
% Radio Polar
c = EjeA^2/EjeB;
% Aplanamiento
% alpha = (a-b)/b;
% Eliminacion del retranqueo  para todos los casos X.
X = X - 500000;
% Elimincacionde retranqueo en Y.
if (strcmp(Hemisferio,'Norte')|| strcmp(Hemisferio,'norte'))
    % Si las coordenadas pertenecen al hemisferio Norte
    % Y no se modifica.
    Y = Y;
else
    % Si las coordenadas pertencen al hemisferio sur.
    % Y = Y -10 000 000;
    Y = Y -10000000;
end
lambdaCent = huso*6 - 183;
% Ecuaciones de Coticchica-Surace
phi = Y /(6366197.724*0.9996);
v = c * 0.9996/sqrt(1+eprima^2*(cos(phi))^2);
a = X/v;
A1 = sin(2*phi);
A2 = A1*(cos(phi))^2;
J2 = phi + A1/2;
J4 = (3*J2 + A2)/4;
J6 = (5*J4 + A2*(cos(phi))^2)/3;
alpha = 0.75*eprima^2;
beta = 5*alpha^2/3;
gama = 35*alpha^3/27;
Bphi = 0.9996*c*(phi - alpha*J2 + beta*J4 -gama*J6);
b = (Y - Bphi)/v;
epsil = (eprima)^2*a^2*(cos(phi))^2/2;
epsilon = a*(1 - epsil/3);
n = b*(1- epsil) + phi;
deltalamb = atan(sinh(epsilon)/cos(n));
tao = atan(cos(deltalamb)*tan(n));
Longitud = deltalamb*180/pi + lambdaCent;
Latitud = (phi + (1 + eprima^2*(cos(phi))^2-1.5*eprima^2*sin(phi)*cos(phi)*(tao-phi))*(tao-phi))*180/pi;
end
```

Función obtención de Bordes de una imagen segmentada

```
function [ im1 ] = ObtenerBorde( imb1 )
imb1 = L;imBin = imb1(:,:,1)>0;
% Delimitar Bordes de la Region con Operadores Morfologicos
% Dilatar y Erosionar con elemento Structurante.
se = strel('octagon',3);
% DILATAR: Numero de veces dilatar N
N = 1;
for i=1:N
    imDila = imdilate(imBin,se);
end
```

```matlab
% EROSIONAR: Numero de veces a erosionar N
N = 1;
imEros=imDila;
for i=1:N
    imEros = imerode(imEros,se);
end
imBorde = imDila -imEros;
imagen = imagenLago;
im1 = imBorde;
```

Función Obtención de Perímetro

```matlab
function [ Perimetro, Area ] =CalcularPerimetro( Imagen )
%CalcularPerimetro: Calcula el perimetro y area de una imagen borde
[m,n]=size(Imagen);
Im=ones(m,n)-Imagen;
Ima=zeros(m,n);

k=find(Imagen>0);
Area=length(k);

masc=[0,1,0;1,0,1;0,1,0];
A=zeros(3);
for i=2:(m-1)
    for j=2:(n-1)
        if Im(i,j)==0
            Ima(i,j)=1;
            A=Im(i-1:i+1,j-1:j+1);
            B=masc&A;
            if sum(sum(B))==0
                Ima(i,j)=0;
            end

        end
    end
end

ImPeri=Ima;
pLado=0;
pDiag=0;
for i=2:(m-1)
    for j=2:(n-1)
        if Ima(i,j)==1
            if Ima(i,j+1) ||Ima(i+1,j)
                pLado=pLado+1;
            end
            if Ima(i+1,j-1) ||Ima(i+1,j+1)
                pDiag=pDiag+1;
            end
        end
    end
end
%Perimetro=pLado+1.41*pDiag;
Perimetro=pLado*0.030+1.41*pDiag*0.030;

disp(Perimetro);
end
```